%% file: tobar_DSP2017.tex
\documentclass[conference]{IEEEtran}
\ifCLASSINFOpdf
\else
\fi
\usepackage{amsmath}
\usepackage{amssymb}
\usepackage{graphicx}

\usepackage{epstopdf}
\usepackage{subfig}
\usepackage[font=small,labelfont=bf]{caption}

\usepackage[utf8]{inputenc}
\usepackage{color}

\newcommand{\cH}{\mathcal{H}}
\newcommand{\cD}{\mathcal{D}}
\newcommand{\x}{\mathbf{x}}

\newcommand{\s}{\mathbf{s}}
\newcommand{\vx}{\vec{\mathbf{x}}}
\newcommand{\vs}{\vec{\mathbf{s}}}
\newcommand{\SE}{K_\text{G}}
\newcommand{\NSE}{K_\text{UG}}
\newcommand{\N}{\mathbb{N}}

\begin{document}
\title{Improving Sparsity in Kernel Adaptive Filters \\Using a Unit-Norm Dictionary}

\author{
\IEEEauthorblockN{Felipe Tobar}
\IEEEauthorblockA{Center for Mathematical Modeling\\
Universidad de Chile\\
Email: \texttt{ftobar@dim.uchile.cl}}
}

\maketitle

\IEEEpeerreviewmaketitle

\input{abstract}
\input{intro}

\input{background}
\input{proposal}
\input{experiments}

\input{conclusions}

\section*{Acknowledgments}
This work was partially supported by Conicyt projects PAI-82140061 and Basal-CMM.

\bibliographystyle{ieeetr}
\bibliography{library}

\end{document}

%% file: abstract.tex

\begin{abstract}
Kernel adaptive filters, a class of adaptive nonlinear time-series models, are known by their ability to learn expressive autoregressive patterns from sequential data. However, for trivial monotonic signals, they struggle to perform accurate predictions and at the same time keep computational complexity within desired boundaries. This is because new observations are incorporated to the dictionary when they are far from what the algorithm has seen in the past. We propose a novel approach to kernel adaptive filtering that compares new observations against dictionary samples in terms of their unit-norm (normalised) versions, meaning that new observations that \emph{look like} previous samples but have a different magnitude are not added to the dictionary. We achieve this by proposing the unit-norm Gaussian kernel and define a sparsification criterion for this novel kernel. This new methodology is validated on two real-world datasets against standard KAF in terms of the normalised mean square error and the  dictionary size.
\end{abstract}

%% file: intro.tex

\section{Introduction}

Support vector machines (SVM) \cite{scholkopf01} are known to be a competitive, intuitive and theoretically-grounded alternative to neural networks within the Machine Learning community. Although SVM were originally envisioned to be trained online using batches of data,  novel kernel methods for regression have been designed in the last decade to learn sequentially, this class is termed kernel adaptive filters (KAFs) \cite{liu2010}. The construction of KAFs  has been possible by combining the properties of representation on high-dimensional feature spaces \cite{steinwart01} together with adaptive learning approaches from the Signal Processing community, using e.g. the least-mean-square (LMS) or recursive-least-square (RLS) rationales \cite{haykin08}. Due to the simplicity of their implementation and intuitive presentation, KAFs have been used in a number of applications from medicine \cite{georga16} to telecommunications \cite{van2006sliding}; moreover, KAF is an active field of research in terms of kernel design \cite{tobar_quat2,tobar_quat1,paul_quat}, automatic determination of model orders \cite{Zhao2016}, and learning approaches \cite{tobar15a}.

In the same manner that SVM and similarity-based modelling \cite{tobar_sim} operate, KAF predicts values of a time-series assessing the similarity between the observations of the signal and a dictionary of historical input-output data. This notion of similarity, given by the kernel function, is rarely studied in practice, where the standard Gaussian kernel is the \textit{de facto} alternative, and the similarity between observations and the dictionary is measured in terms of the Euclidean distance. One drawback of this choice of similarity is that when the signal moves to a region that is unknown for the dictionary, the KAF prediction reverts to zero even when it is clear that the signal cannot take that value. This makes KAF struggle to learn very simple signals, for instance, a linear function would require a monotonically-increasing number of dictionary members, since the signal is always increasing in magnitude and therefore moving away from the current dictionary. 

We address this issue by using a unit-norm representation of the input to the kernel, meaning that the similarity between observations and dictionary centres is calculated as the Euclidean distance of their normalised (unit-norm) versions, or in other words, they are compared \textit{up to their magnitude}. Our hypothesis is that this will help us to acquire knowledge from one region and then extrapolate this knowledge to regions where the signal looks similar but with a different magnitude. We achieve this by proposing a novel kernel, the unit-norm Gaussian kernel, that retains the useful properties of the standard Gaussian kernel but at the same time incorporates the unit-norm comparison. We also describe a specific sparsification criteria for our model, which builds the dictionary from unit-norm data, this allows us to derive a simple and intuitive form of the proposed kernel predictor. Through illustrative examples using both synthetic and real-world datasets we show the drawbacks of current KAF methods and how these are sidestepped by the proposed unit-norm formulation of KAF.

%% file: background.tex

\section{Kernel Adaptive Filters }

Kernel adaptive filters \cite{liu2010} are a class of autoregressive nonlinear models for time series that operate by embedding observations of the signal onto a feature space; then, the feature embeddings are combined with parameters adapted online to produce the prediction. The closed form of the KAF predictor is possible due to the use of a reproducing kernel Hilbert space (RKHS) \cite{scholkopf01} as feature space, this allows us to express inner products among features as kernel evaluations. 

\subsection{Model specification} 
\label{sub:model_specification}

Let us define a time series by $\{y_i\}_{i\in\N}$, denote the order of the filter by $d$ and the input to the filter at time $i$ by $\x_i = [y_{i-d},\ldots,y_{i-1}]^\top$. The first step of a KAF is to map $\x_i$ onto an RKHS $\cH$ through $\phi_{\x_i}$ and then approximate $y_i$ linearly as

\begin{equation}
	\hat{y}_i = \langle \phi_{\x_i} , W \rangle
\end{equation}
where $\langle \cdot, \cdot \rangle$ is the inner product in $\cH$ and the feature weight $W\in\cH$ is chosen according to the Representer Theorem \cite{scholkopf_representer}, which states that the weight $W$ that minimises any square loss is a linear combination of the feature samples $\phi_{\x_i}$, where the $\x_i$ are observations. In the online operation case, this corresponds to considering the entire history of the signal up to time $i$, that is, denoting the new weights $\alpha_{i,j}$, the optimal weight at time $i$ has the form $W_i = \sum_{j=1}^{i-1}\alpha_{i,j}\phi_{\x_j}$. Additionally, as the inner product between feature samples can be expressed as kernel evaluations \cite{scholkopf01}, we have
\begin{align}
	\hat{y}_i &= \langle \phi_{\x_i} , \sum_{j=1}^{i-1}\alpha_{i,j}\phi_{\x_j} \rangle\\
	&= \sum_{j=1}^{i-1}\alpha_{i,j} \langle \phi_{\x_i} , \phi_{\x_j} \rangle\\
	&= \sum_{j=1}^{i-1}\alpha_{i,j} K(\x_i, \x_j)
\end{align}
where the parameters of the model $\alpha_{i,j}$ can be adapted online using e.g., LMS \cite{liu2010} or RLS \cite{engel04}. In this form, the KAF estimator is unsuitable for online operation due to the fact that the number of terms in the above summation grows unbounded, however, we can consider the so-called sparsification criteria that selects a subset of the regressors $\x_i$ to produce sound estimates and keep computational complexity at bay. 

\subsection{Sparsification criteria} 
\label{sub:sparsification_criteria}
There are three main sparsification criteria used in KAF. First, the novelty criterion \cite{platt91}, where a new sample is added to the dictionary only if it is far enough from the dictionary and if the prediction error associated to that sample was large. Second, the approximate linear dependence (ALD) criterion \cite{engel04}, which includes a new member in the dictionary if the feature sample associated to such element is \emph{approximately linearly dependent} of the feature dictionary members in the RKHS. Third, the coherence criterion \cite{richard09} which is a simplified variant of ALD that only considers the distance between the input and the closest dictionary member instead of the entire dictionary. All sparsification criteria produce a dictionary in time against which the new input is compared (instead of using all historical observations), with this choice, the new form of the KAF predictor is 
\begin{align}
	\hat{y}_i &= \sum_{j=1}^{i-1}\alpha_{i,j} K(\x_i, \s_j)
\end{align}
where the set $\cD_i = \{\s_j\}_{N_i}$ is known as dictionary at time $i$ and the vectors $\s_j$ as centres. 

Sparsification criteria for KAFs ensure finite-sized dictionary when the signal lies on a compact set \cite{honeine15}, however, when the signal is monotonically increasing these criteria keep including more centres in the dictionary, thus resulting in higher computational complexity due to having dictionary members that do not necessarily improve predictions.

\subsection{Illustrative example using synthetic data} 
\label{sub:example}

\begin{figure}[h!]
	\centering
	\includegraphics[width=0.5\textwidth]{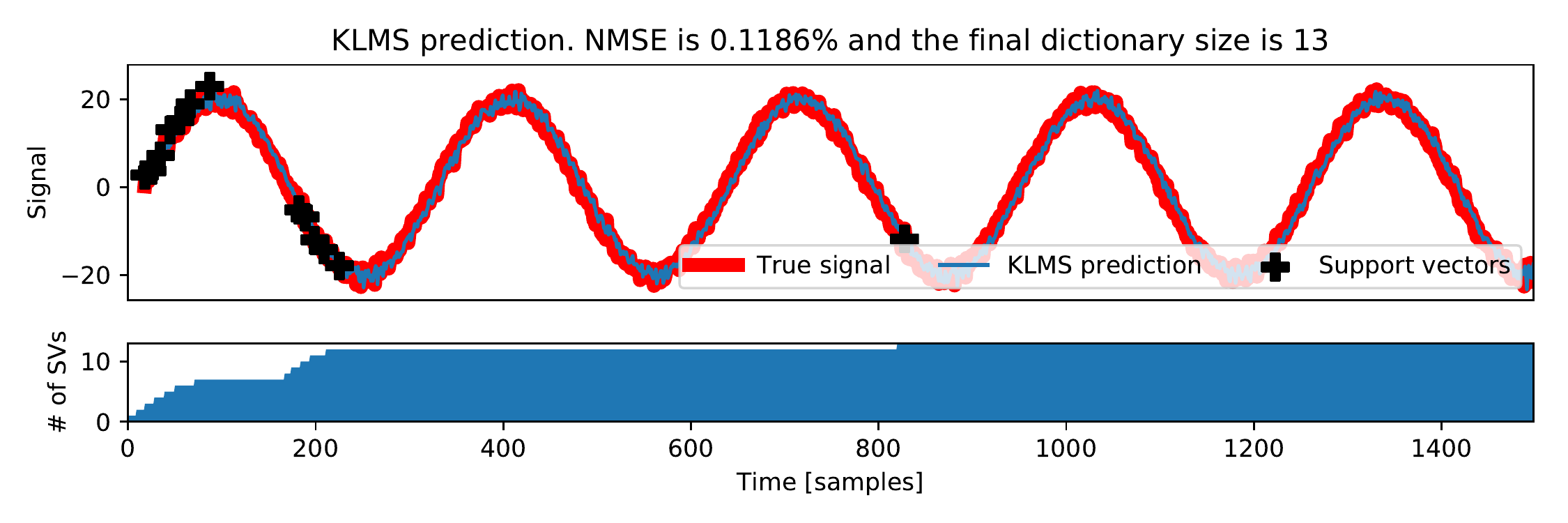} 
	\caption{KLMS prediction for a synthetic sinewave: Top plot shows the true and predicted signals together with the centres added and the bottom plot the accumulated number centres in time}
	\label{fig:KLMS_sine}
\end{figure}

We considered a synthetic sinewave and implemented a kernel least mean square algorithm \cite{liu08} using the novelty sparsification criterion, the motivation for this example was to train KLMS on data that lie on a compact set and therefore see that the dictionary remains bounded. The result of the implementation is shown in Fig. \ref{fig:KLMS_sine}, in the top plot the true sinewave signal (red) and the KLMS prediction (blue) are shown, together with the values of the signal that required the addition of dictionary elements in black crosses; the bottom plot shows the number of support vectors accumulated in time. From this figure we can see that the aim of the sparsification criterion chosen is fulfilled: the addition of samples to the dictionary occurs in the initial part of the experiment since the range of the data is remains fixed in time .  

%% file: proposal.tex

\section{KAF with Unit-Norm Dictionary Centres} 
\label{sec:a_novel_kernel_adaptive_filter}

Our aim is to overcome the unbounded growth of the dictionary of kernel adaptive filters when the range of the signal expands in time. To this end, we will populate the dictionary with normalised (unit-norm) versions of the observed samples, this way, the dictionary will not contain centres that are close-to-proportional to one another, but it will only contain centres that differ in their direction rather than magnitude. Furthermore, as the dictionary will be \emph{blind} to the magnitude of the data, we will preserve the input magnitude by amplifying the kernel evaluation by the magnitude of the input. We present our approach by proposing a novel kernel as follows.

\subsection{The unit-norm Gaussian kernel} 
 \label{sub:the_unit_norm_gaussian_kernel}
 
A standard kernel within KAF is the Gaussian kernel defined by 
 \begin{equation}
 	\SE(\x_1,\x_2) = \exp\left(\frac{-1}{2l^2} ||\x_1-\x_2||^2 \right)
 	\label{eq:SE}
 \end{equation}
 where $||\cdot||$ denotes the norm in the input space and $l>0$ is referred to as the kernel lengthscale. 

 We propose a novel kernel termed unit-norm Gaussian kernel (UG) defined as follows
  \begin{equation}
 	\NSE(\x_1,\x_2) = ||\x_1||\exp\left(\frac{-1}{2l^2} ||\vx_1-\vx_2||^2 \right)||\x_2||
	\label{eq:NSE}
 \end{equation}
 where  $\vx = \x/||\x||$ is the unit-norm (normalised) version of $\x$ that preserves the direction of $\x$ but not its magnitude. Let us see that the proposed $\NSE$ kernel is a valid positive-definite kernel, in effect, denoting by $\phi$ the eigenfunction of the $\SE$ kernel, i.e., $\SE(\x_1,\x_2) = \langle \phi_{\x_1}, \phi_{\x_2}\rangle$, the proposed $\NSE$ can be decomposed into the inner product form
 \begin{align}
 	\NSE(\x_1,\x_2) &=||\x_1||\SE(\vx_1,\vx_2)||\x_2|| \\
 	&=||\x_1|| \langle \phi_{\vx_1}, \phi_{\vx_2}\rangle\x_2|| \nonumber\\
 	&= \langle ||\x_1||\phi_{\vx_1}, ||\x_2||\phi_{\vx_2} \rangle\nonumber\\
 	&= \langle \psi_{\x_1}, \psi_{\x_2} \rangle\nonumber
 \end{align}
 where $\psi_{\x} = ||\x||\phi_{\vx}$ is the feature expansion of the proposed $\NSE$ kernel. Therefore, as $\NSE$ admits the above inner product decomposition, it is a positive definite kernel \cite{scholkopf01}.

 Besides the fact that the proposed kernel aims to avoid redundancies in the dictionary by only considering unit-norm versions of the inputs, $\NSE$ in eq. \eqref{eq:NSE} poses a clear advantage in parameter setting. In the standard Gaussian kernel in eq. \eqref{eq:SE} the lengthscale parameter of the kernel is set based on both the dimension and magnitude of the input samples, meaning that for each new set of data, the lengthscale has to be hand-tuned using heuristics or computationally-demanding methods \cite{dsp2017a}. Conversely, as all the inputs to the proposed $\NSE$ kernel are unit-norm, the lengthscale parameter can be chosen based only on the dimension, in fact, by assuming that the difference $\vx_1-\vx_2$ in eq. \eqref{eq:NSE} is isotropic, the lengthscale parameter $l$ can be set to $l=l_0\sqrt{d}$, where $d$ is the dimension of the input and $l_0$ is the \textit{lengthscale per coordinate}. Experimental evaluations revealed that values for $l_0$ between 1 and 3 are suitable options for a wide range of data sets when using the proposed kernel $\NSE$.  


 \subsection{KLMS with the unit-norm Gaussian kernel} 
 \label{sub:klms_with_the_unit_norm_gaussian_kernel}

Taking advantage of the relationship between $\SE$ and $\NSE$ in eqs. \eqref{eq:SE}-\eqref{eq:NSE}, the kernel prediction of a time series using the unit-norm Gaussian kernel can be expressed in terms of the standard Gaussian kernel as 
 \begin{align}
 	\hat{y}_i &= \sum_{j=1}^{N_i}\alpha_{i,j}\NSE(\x_i,\s_j)\\
 	&= \sum_{j=1}^{N_i}\alpha_{i,j}||\x_i||\SE(\vx_i,\vs_j)||\s_j||\nonumber
 \end{align}
 Furthermore, if we only record the unit-norm versions of the input samples in the dictionary, the above expression can be further simplified into 
 \begin{align}
 	\hat{y}_i = \sum_{j=1}^{N_i}\alpha_{i,j}||\x_i||\SE(\vx_i,\vs_j)
 	\label{eq:unKLMS}
 \end{align} 
 where recall that  $\forall j\ \vs_j = \s_j$ since they are unit-norm.

 Choosing the LMS update rule for the weights $\{\alpha_{i,j}\}_{i\in\N, j=1:N_i}$ (KLMS), that is, denoting the prediction error at time $i$ as $e_i = y_i - \hat{y}_i$, the update rule of the $j^\text{th}$ weight, with learning rate $\mu_0$, is given by 
 \begin{align}
 \alpha_{i+1,j} &= \alpha_{i,j} - \frac{\mu_0}{2}\frac{\partial e_i^2}{\partial\alpha_{i,j}}\\
 & =  \alpha_{i,j} - \mu_0 e_i \frac{\partial e_i}{\partial\alpha_{i,j}}\nonumber\\
 & =  \alpha_{i,j} + \mu_0 \frac{\partial }{\partial\alpha_{i,j}}\left(\sum_{j=1}^{N_i}\alpha_{i,j}||\x_i||\SE(\vx_i,\vs_j)\right)\nonumber\\
  & =  \alpha_{i,j} + \mu_0 ||\x_i||\SE(\vx_i,\vs_j).\nonumber
\end{align}

Notice that unlike the KLMS with Gaussian kernel, the sequential correction to the weights depends of the magnitude of the input, this hinders the choice of the learning rate since the optimal choice for $\mu_0$ depends directly from the norm of the data samples and has to be tuned for each dataset---if $\mu_0$ is not chosen carefully, convergence of the KLMS is not guaranteed. As we are precisely interested in cases where the signal grows unbounded, we use the normalised version of KLMS instead, this is achieved by choosing the learning rate to be inversely proportional to the square norm of the input in eq. \eqref{eq:unKLMS}, that is, 

\begin{equation}
 \mu_0 = \mu\left(\epsilon + ||\x||^{2}\sum_{i=1}^{N_t}\SE^2(\x,\vs_i)\right)^{-1}
\end{equation}
where $\epsilon>0$ is a constant that prevents the above quantity from going to infinity. Setting the new learning rate $\mu$ is now straightforward, since from the LMS convergence properties we know that  convergence is guaranteed for $\mu\in(0,1)$. With this choice, the weight update rule is given by
 \begin{align}
 \alpha_{i+1,j} &=   \alpha_{i,j} + \mu \frac{||\x_i||\SE(\vx_i,\vs_j)}{\epsilon + ||\x_i||^2\sum_{j=1}^{N_i}\SE^2(\x_i,\vs_j)},\ \mu\in (0,1) 
\end{align}
Observe that, as $\epsilon$ is chosen to be as close to zero as possible, the correction term of the weight update of the proposed normalised KLMS with unit-norm Gaussian kernel resembles that of the standard Gaussian kernel but divided by $||\x_i||$. This has a clear interpretation, since the unit-norm Gaussian kernel evaluation in eq. \eqref{eq:NSE} is inversely proportional to the norm of the input and therefore the increments need to be bounded so the estimate does not diverge. From now on we will refer to the proposed approach as the unit-norm KLMS.


 \subsection{Sparisification} 
 \label{sub:sparisification}
 One of the differences between the proposed unit-norm Gaussian kernel in eq. \eqref{eq:NSE} and the standard one in eq. \eqref{eq:SE} is that the latter is a measure of similarity, whereas the former is not due to the proportionality to the norm of the input. For this reason, we consider the novelty sparsification criteria for unit-norm KLMS expressed in terms of the standard Gaussian kernel instead of the Euclidean distance between the unit-norm samples. Specifically, a new input sample $\x$ with unit-norm version $\vx=\x/||\x||$ and output $y$ is added to the dictionary if (i) it is far enough from the dictionary:
 \begin{align}
 	\SE(\vx,\cD_i) = \max_{\vs_j\in\cD_i} \ \SE(\vx,\vs) < \delta_\text{dict}
 \end{align}
 and (ii) if the relative error associated to the prediction of $y$ given the input $\x$ is large enough: 
 \begin{align}
 	|e|/|y| = \frac{1}{|y|}\left|y-\sum_{j=1}^{N_i}\alpha_{i,j}\NSE(\x,\vs_j)\right| > \delta_\text{pred}.
 \end{align}
The reason to compute distance to the dictionary using the Gaussian kernel is that the parameter $\delta_\text{dict}$ is between 0 and 1, and that comparison with the standard Gaussian-kernel KLMS is direct. 

%% file: experiments.tex

\section{Simulations}

We validated the proposed unit-norm KLMS against the standard KLMS in the prediction of two real-world time series. The first experiment considers the sunspots time series, where the aim was to show that the unit-norm KLMS behaves well for compact-range signals just as the standard KLMS. The second experiment studied the Mauna Loa CO$_2$ concentration time series, where the size of the dictionary of the standard KLMS grows unbounded, whereas that of the proposed unit-norm KLMS does not. Both time series are available from  Python's \texttt{statsmodel} datasets.

\subsection{Sunspots time series} 
\label{sub:sunspots}

\begin{figure}[h!]
	\centering
	\includegraphics[width=0.5\textwidth]{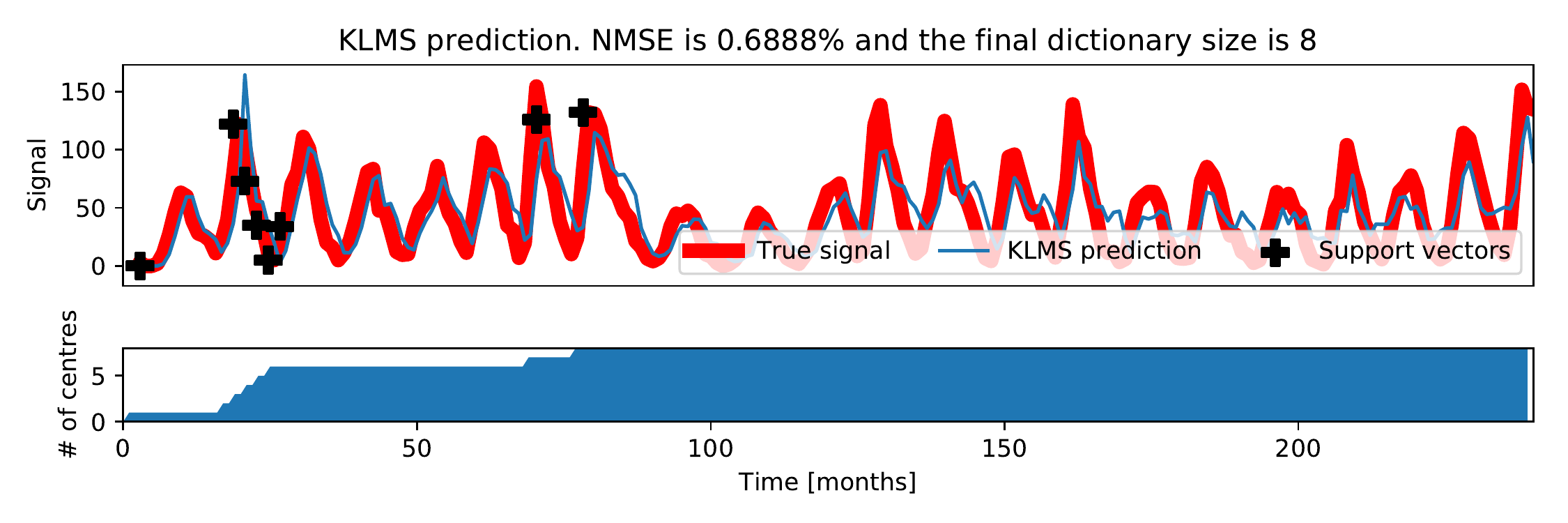}
	\caption{KLMS prediction for the sunspots time series: Top plot shows the true and predicted signals together with the support vectors added and the bottom plot the accumulated number of support vectors in time}
	\label{fig:KLMS_sunspots}
\end{figure}

\begin{figure}[h!]
	\centering
	\includegraphics[width=0.5\textwidth]{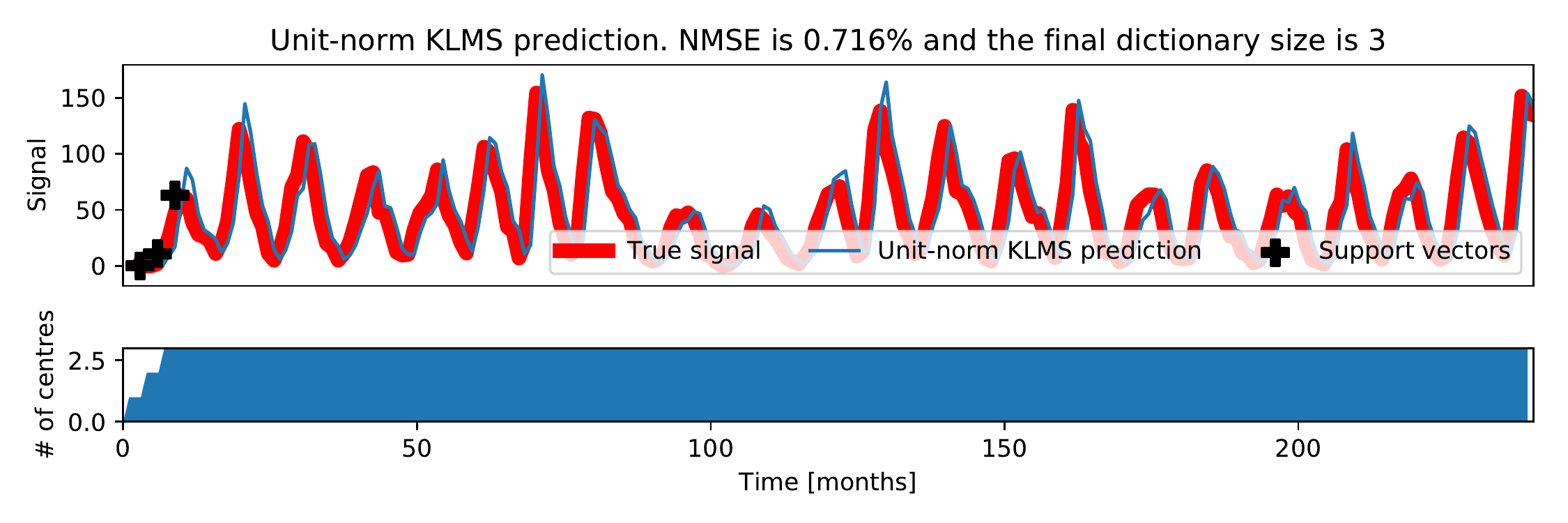}
	\caption{Unit-norm KLMS prediction for the sunspots time series: Top plot shows the true and predicted signals together with the support vectors added and the bottom plot the accumulated number of support vectors in time}
	\label{fig:PKLMS_sunspots}
\end{figure}

We considered 250 samples of the yearly sunspot time series and implemented both the proposed unit-norm and standard KLMS algorithms with the novelty criterion described in the previous sections. The results are shown in Figs. \ref{fig:KLMS_sunspots} for KLMS and \ref{fig:PKLMS_sunspots} for the unit-norm KLMS. For both figures, the title shows the normalised mean square error (NMSE) and the final dictionary size, and the top plot shows the true sunspots signal (thick red line) and the KLMS predictions (thin blue line), together with the values of the signal that required the addition of dictionary centres in black crosses. The bottom plots in both figures show the number of dictionary centres accumulated in time. We can see how the sparsification criteria worked for both kernel predictors by only including centres durign the initial parts of the experiment, since the range of the signal is compact. The peaks of the estimate for the unit-norm KLMS can be attributed to the fact that the proposed algorithm does not consider the magnitude of the input and therefore the prediction tends to grow unless it is learnt otherwise, this resulted in a slightly higher NMSE for the proposed model.

\subsection{Mauna Loa CO$_2$ time series} 
\label{sub:maunaloa}

\begin{figure}[h!]
	\centering
	\includegraphics[width=0.5\textwidth]{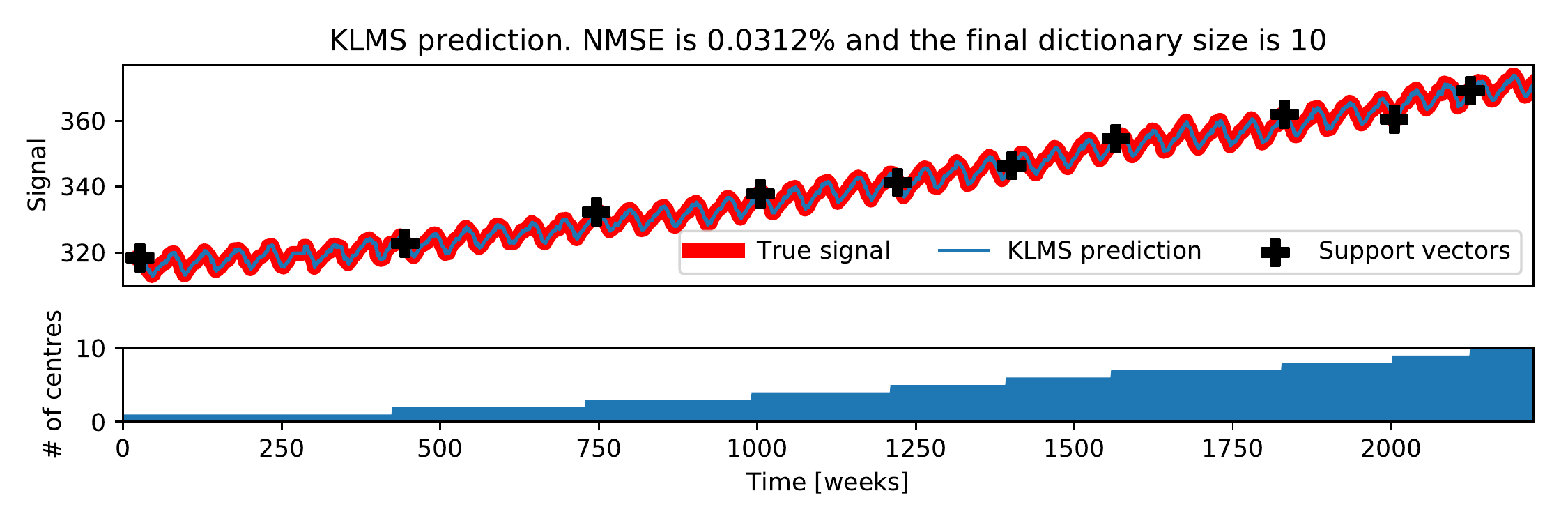} 
	\caption{Standard KLMS prediction for the Mauna Loa CO$_2$ time series: Top plot shows the true and predicted signals together with the support vectors added and the bottom plot the accumulated number of support vectors in time}
	\label{fig:KLMS_maunaloa}
\end{figure}

The Mauna Loa dataset contains the weekly concentration of CO$_2$  collected at the Mauna Loa Observatory in Hawaii, between 1958 and 2001; the series has missing values, which we replaced by the previous value for this experiment. This signal has both increasing and semi-periodic components. From Fig. \ref{fig:KLMS_maunaloa} we can see how the number of dictionary centres grows linearly with the range of the signal when using the standard KLMS, this evidences the drawback of the KLMS approach for increasing time series, even though the signal follows a fairly constant semi-periodic pattern with increasing trend. 

\begin{figure}[h!]
	\centering
	\includegraphics[width=0.5\textwidth]{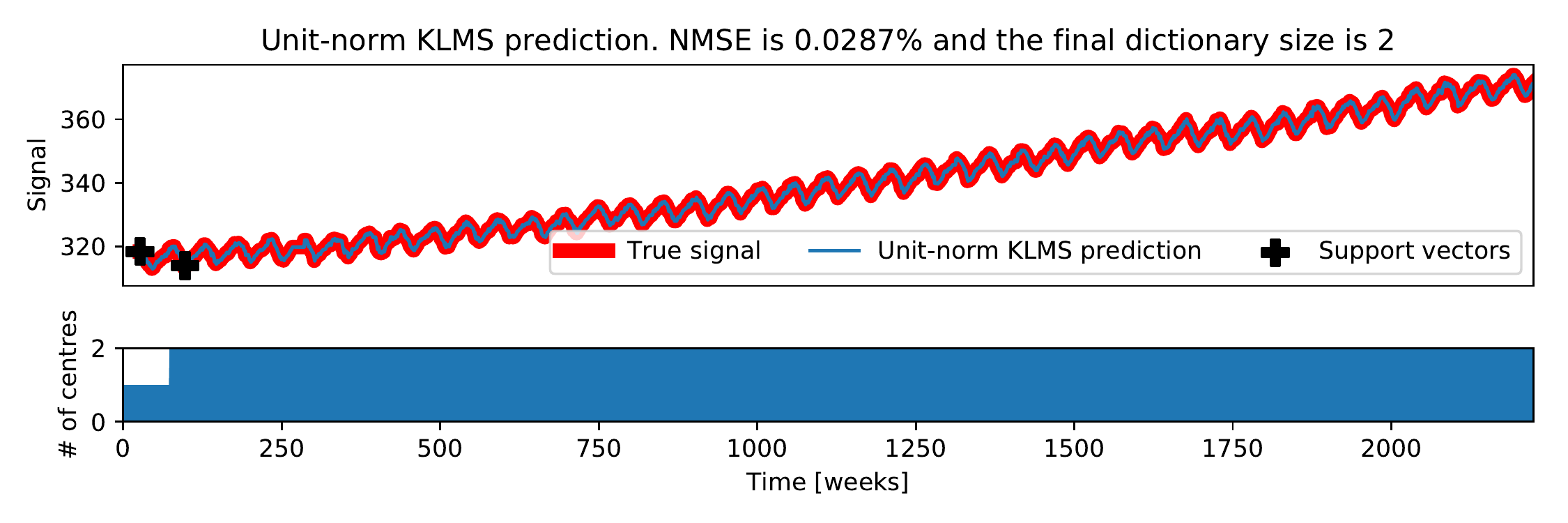} 
	\caption{Unit-norm KLMS prediction for the Mauna Loa CO$_2$ time series: Top plot shows the true and predicted signals together with the support vectors added and the bottom plot the accumulated number of support vectors in time}
	\label{fig:PKLMS_maunaloa}
\end{figure}

\begin{figure}[h!]
	\centering
	\includegraphics[width=0.5\textwidth]{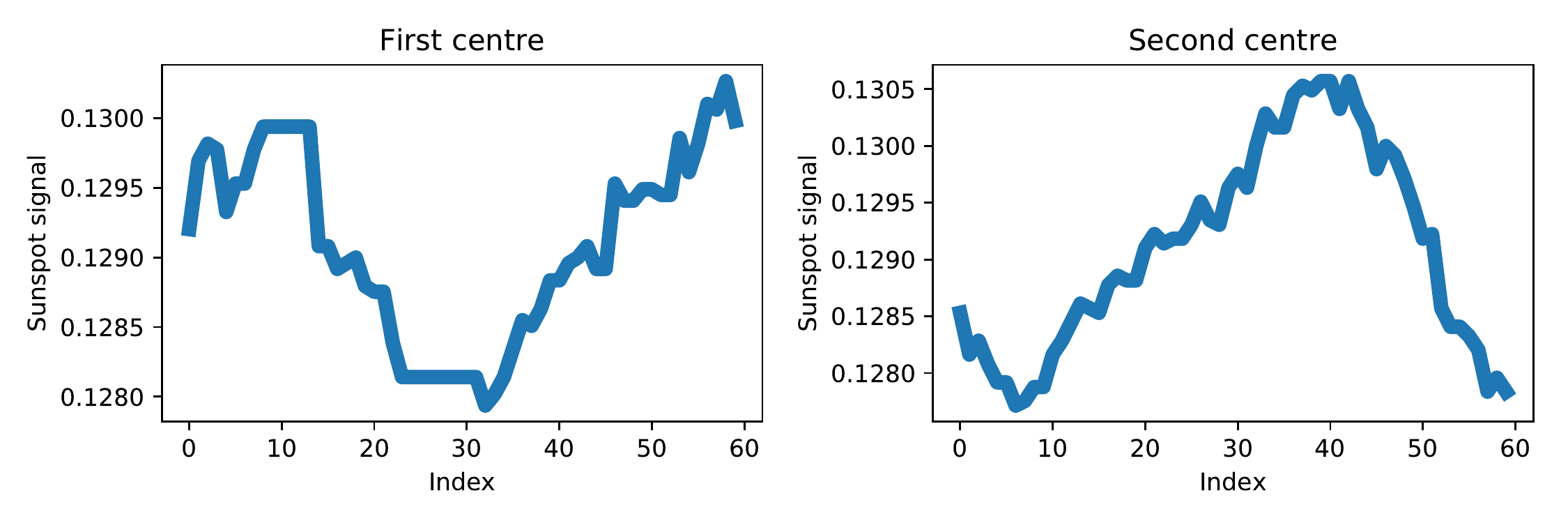} 
	\caption{Dictionary centres of unit-norm KLMS for the Mauna Loa CO$_2$ time series. These elements represents the information encapsulated by the proposed unit-norm KLMS, where the algorithm only requires to assess whether the signal is in the concave or convex part of the cycle, but not its magnitude, to compute the prediction.}
	\label{fig:centres}
\end{figure}

Conversely, the proposed unit-norm KLMS predictor only 
required two dictionary centres to predict the Mauna Loa time series with high accuracy, thus representing an improvement over the standard KLMS in terms of the normalised mean square error---see Fig. \ref{fig:PKLMS_maunaloa}. An interesting feature of the unit-norm KLMS prediction of the Mauna Loa dataset can be appreciated analysing the dictionary: Fig. \ref{fig:centres} shows the two centres considered by the algorithm, this suggests that the algorithm evaluates similarity between the input and the convex and concave parts of the signal to perform the prediction. This revealed that the algorithm learnt the repetitive behaviour of the series owing to the unit-norm representation, since the standard KLMS does not recognise repetitive patterns if they are of different magnitude. 

%% file: conclusions.tex

\section{Conclusions}

To overcome the increased computational complexity related to learn monotonic time series using kernel adaptive filters, we have proposed the unit-norm Gaussian kernel and show how it can be used within a kernel least mean square setting. The proposed approach only assess similarity between observed inputs and the dictionary in terms of their direction and not magnitude, thus avoiding redundancies among samples that have similar shape but different magnitude. We have illustrated the shortcomings of standard KAFs that include observations in the dictionary based on the Euclidean norm only, and have validated our proposed approach to sidestep this drawback. Through experimental validation using real-world data, we have confirmed that the unit-norm approach to KAF outperforms standard methods in terms of dictionary size (model complexity) and normalised mean square error (predictive ability).

%% file: tobar_DSP2017.bbl
\begin{thebibliography}{10}

\bibitem{scholkopf01}
B.~Scholkopf and A.~J. Smola, {\em {Learning with Kernels: Support Vector
  Machines, Regularization, Optimization, and Beyond}}.
\newblock MIT Press, 2001.

\bibitem{liu2010}
W.~Liu, J.~C. Principe, and S.~Haykin, {\em Kernel Adaptive Filtering: A
  Comprehensive Introduction}.
\newblock Wiley, 2010.

\bibitem{steinwart01}
I.~Steinwart, ``On the influence of the kernel on the consistency of support
  vector machines,'' {\em J. Mach. Learn. Res.}, vol.~2, pp.~67--93, 2001.

\bibitem{haykin08}
S.~S. Haykin, {\em Adaptive filter theory}.
\newblock Pearson Education India, 2008.

\bibitem{georga16}
E.~I. Georga, J.~C. Principe, D.~Polyzos, and D.~I. Fotiadis, ``Non-linear
  dynamic modeling of glucose in type 1 diabetes with kernel adaptive
  filters,'' in {\em Proc. of the IEEE International Conference of Engineering
  in Medicine and Biology Society (EMBC)}, pp.~5897--5900, Aug 2016.

\bibitem{van2006sliding}
S.~Van~Vaerenbergh, J.~Via, and I.~Santamar{\'\i}a, ``A sliding-window kernel
  rls algorithm and its application to nonlinear channel identification,'' in
  {\em Proc. of ICASSP}, vol.~5, pp.~V--V, IEEE, 2006.

\bibitem{tobar_quat2}
F.~Tobar and D.~P. Mandic, ``Design of positive-definite quaternion kernels,''
  {\em IEEE Signal Processing Letters}, vol.~22, pp.~2117--2121, Nov 2015.

\bibitem{tobar_quat1}
F.~A. Tobar and D.~P. Mandic, ``Quaternion reproducing kernel hilbert spaces:
  Existence and uniqueness conditions,'' {\em IEEE Transactions on Information
  Theory}, vol.~60, pp.~5736--5749, Sept 2014.

\bibitem{paul_quat}
T.~K. Paul and T.~Ogunfunmi, ``A kernel adaptive algorithm for
  quaternion-valued inputs,'' {\em IEEE Transactions on Neural Networks and
  Learning Systems}, vol.~26, pp.~2422--2439, Oct 2015.

\bibitem{Zhao2016}
S.~Zhao, B.~Chen, Z.~Cao, P.~Zhu, and J.~C. Principe, ``Self-organizing kernel
  adaptive filtering,'' {\em EURASIP Journal on Advances in Signal Processing},
  vol.~2016, no.~1, p.~106, 2016.

\bibitem{tobar15a}
F.~Tobar, P.~Djuri\'c, and D.~Mandic, ``Unsupervised state-space modelling
  using reproducing kernels,'' {\em IEEE Trans. on Signal Processing}, pp.~5210
  -- 5221, 2015.

\bibitem{tobar_sim}
F.~A. Tobar, L.~Yacher, R.~Paredes, and M.~E. Orchard, ``Anomaly detection in
  power generation plants using similarity-based modeling and multivariate
  analysis,'' in {\em Proc. of the American Control Conference},
  pp.~1940--1945, June 2011.

\bibitem{scholkopf_representer}
B.~Sch{\"o}lkopf, R.~Herbrich, and A.~J. Smola, {\em A Generalized Representer
  Theorem}, pp.~416--426.
\newblock Berlin, Heidelberg: Springer, 2001.

\bibitem{engel04}
Y.~Engel, S.~Mannor, and R.~Meir, ``{The kernel recursive least-squares
  algorithm},'' {\em IEEE Trans. on Signal Processing}, vol.~52, no.~8,
  pp.~2275--2285, 2004.

\bibitem{platt91}
J.~Platt, ``A resource-allocating network for function interpolation,'' {\em
  Neural Computation}, vol.~3, no.~2, pp.~213--225, 1991.

\bibitem{richard09}
C.~Richard, J.~Bermudez, and P.~Honeine, ``Online prediction of time series
  data with kernels,'' {\em IEEE Trans. on Signal Processing}, vol.~57, no.~3,
  pp.~1058 --1067, 2009.

\bibitem{honeine15}
P.~Honeine, ``Analyzing sparse dictionaries for online learning with kernels,''
  {\em IEEE Trans. on Signal Processing}, vol.~63, no.~23, pp.~6343--6353,
  2015.

\bibitem{liu08}
W.~Liu, P.~Pokharel, and J.~Principe, ``The kernel least-mean-square
  algorithm,'' {\em IEEE Trans. on Signal Processing}, vol.~56, no.~2, pp.~543
  --554, 2008.

\bibitem{dsp2017a}
I.~Castro, C.~Silva, and F.~Tobar, ``Initialising kernel adaptive filters via
  probabilistic inference,'' in {\em Proc. of the Digital Signal Processing
  conference.}, 2017.

\end{thebibliography}
